\documentclass[acmtog,nonacm]{acmart}

\usepackage{booktabs} 

\usepackage[ruled]{algorithm2e} 

\SetAlFnt{\small}
\SetAlCapFnt{\small}
\SetAlCapNameFnt{\small}
\SetAlCapHSkip{0pt}

\acmJournal{TOG}

\newcommand{\ie}{\textit{i}.\textit{e}.~}
\newcommand{\etc}{\textit{etc}~}

\usepackage{graphicx}
\usepackage{amsmath}
\usepackage{booktabs}
\usepackage{color}
\usepackage{colortbl}  
\usepackage{xcolor}
\usepackage{array}
\usepackage{bbding}
\usepackage{colortbl}
\newcommand{\first}{\cellcolor[HTML]{F4B5B4}}
\newcommand{\second}{\cellcolor[HTML]{F9DAB7}}
\newcommand{\third}{\cellcolor[HTML]{FFFFBB}}
\newcommand{\shaddow}{\cellcolor[HTML]{EFEFEF}}

\usepackage{lipsum}

\makeatletter
\renewcommand\@authorsaddresses{${}^*$ Equal contribution.\quad ${}^\dag$ Corresponding author}
\makeatother
\renewcommand\footnotetextcopyrightpermission[1]{}

\begin{document}
\title{\textcolor{purple}{DiMeR}: \textcolor{purple}{Di}sentangled \textcolor{purple}{Me}sh \textcolor{purple}{R}econstruction Model}

\author{Lutao Jiang$^{*}$}
\affiliation{%
 \institution{HKUST(GZ)}
 \country{China}
}
\email{ljiang553@connect.hkust-gz.edu.cn}

\author{Jiantao Lin$^{*}$}
\affiliation{%
 \institution{HKUST(GZ)}
 \country{China}
}
\email{jlin695@connect.hkust-gz.edu.cn}

\author{Kanghao Chen$^{*}$}
\affiliation{%
 \institution{HKUST(GZ)}
 \country{China}
}
\email{kchen879@connect.hkust-gz.edu.cn}

\author{Wenhang Ge$^{*}$}
\affiliation{%
 \institution{HKUST(GZ)}
 \country{China}
}
\email{wge950@connect.hkust-gz.edu.cn}

\author{Xin Yang}
\affiliation{%
 \institution{HKUST(GZ)}
 \country{China}
}
\affiliation{%
 \institution{HKUST}
 \country{China}
}
\email{xyangbk@connect.ust.hk}

\author{Yifan Jiang}
\affiliation{%
 \institution{HKUST(GZ)}
 \country{China}
}
\email{yjiang578@connect.hkust-gz.edu.cn}

\author{Yuanhuiyi Lyu}
\affiliation{%
 \institution{HKUST(GZ)}
 \country{China}
}
\email{ylyu650@connect.hkust-gz.edu.cn}

\author{Xu Zheng}
\affiliation{%
 \institution{HKUST(GZ)}
 \country{China}
}
\email{xzheng287@connect.hkust-gz.edu.cn}

\author{Yinchuan Li}
\affiliation{%
 \institution{Noah's Ark Lab}
 \country{China}
}
\email{yinchuan.li.cn@gmail.com}

\author{Ying-Cong Chen$^{\dag}$}
\affiliation{%
 \institution{HKUST(GZ)}
 \country{China}
}
\affiliation{%
 \institution{HKUST}
 \country{China}
}
\email{yingcongchen@ust.hk}

\renewcommand\shortauthors{Jiang, L. et al}

\begin{teaserfigure}
\centering
\vspace{-10pt}
  \includegraphics[width=1.0\textwidth]{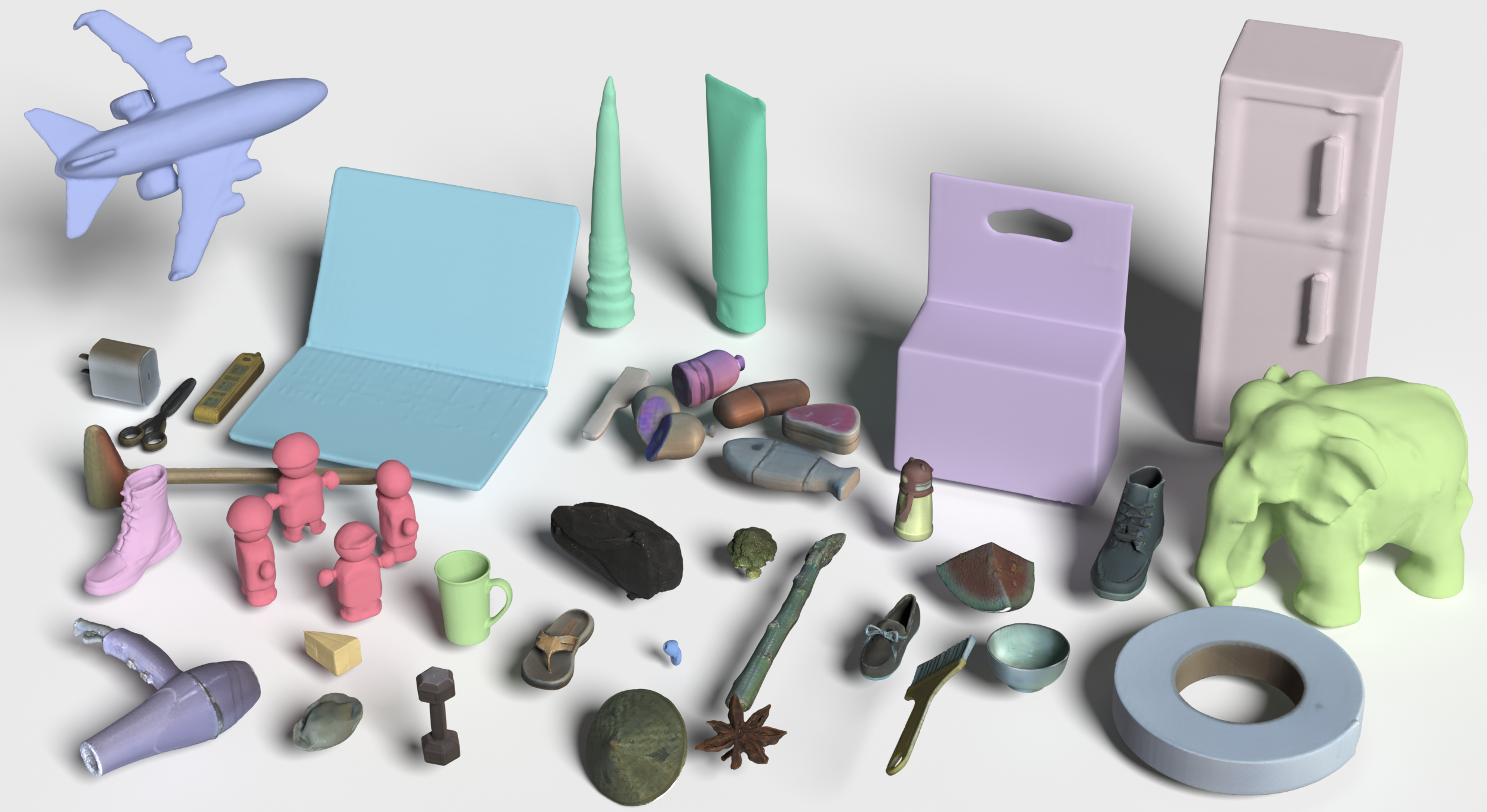}
  \vspace{-20pt}
  \caption{DiMeR takes the image or text inputs and generates detailed 3D meshes.}
  \label{fig: teaser}
\end{teaserfigure}

\begin{abstract}

We propose DiMeR, a novel geometry-texture disentangled feed-forward model with 3D supervision for sparse-view mesh reconstruction.
Existing methods confront two persistent obstacles: (i) textures can conceal geometric errors, \ie, visually plausible images can be rendered even with wrong geometry, producing multiple ambiguous optimization objectives in geometry-texture mixed solution space for similar objects; 
and (ii) prevailing mesh extraction methods are redundant, unstable, and lack 3D supervision.
To solve these challenges, we rethink the inductive bias for mesh reconstruction.
First, we disentangle the unified geometry-texture solution space, where a single input admits multiple feasible solutions, into geometry and texture spaces individually.
Specifically, given that normal maps are strictly consistent with geometry and accurately capture surface variations, the normal maps serve as the sole input for geometry prediction in DiMeR, while the texture is estimated from RGB images.
Second, we streamline the algorithm of mesh extraction by eliminating modules with low performance/cost ratios and redesigning regularization losses with 3D supervision.
Notably, DiMeR still accepts raw RGB images as input by leveraging foundation models for normal prediction.
Extensive experiments demonstrate that DiMeR generalises across sparse‑view-, single‑image-, and text‑to‑3D tasks, consistently outperforming baselines. On the GSO and OmniObject3D datasets, DiMeR significantly reduces Chamfer Distance by more than \textbf{30\%}.
Project Page: \url{https://lutao2021.github.io/DiMeR_page/}
\end{abstract}

\maketitle

\vspace{-10pt}

\section{Introduction}

\begin{figure}[t!]
    \centering
    \includegraphics[width=1\linewidth]{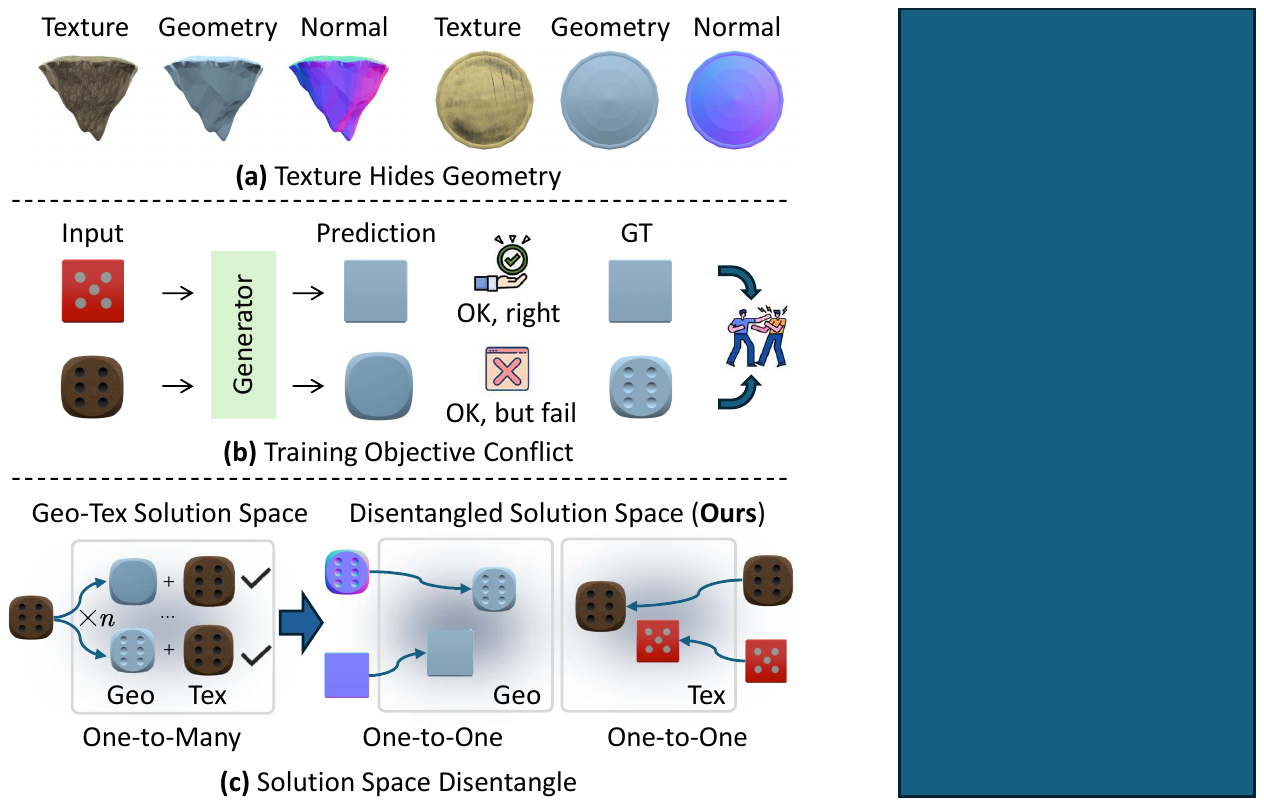}
    \vspace{-20pt}
    \caption{(a) exhibits difficulty in distinguishing geometry from RGB images.
    (b) shows the conflict input-GT pairs in datasets due to problem (a), hindering the training. 
    (c) illustrates our idea: disentangle the mixed solution space containing multiple feasible solutions into two separate spaces with unbiased input.
    Samples are from the Objavers dataset~\cite{deitke2023objaverse}.}
    \vspace{-20pt}
    \label{fig: ambiguity}
\end{figure}

The tasks of 3D reconstruction and generation have garnered significant attention, largely due to the advancements made by NeRF \cite{mildenhall2021nerf} and 3DGS~\cite{kerbl20233d}.
However, transforming them into the mesh poses a challenge.
In this paper, we focus on mesh representation, which is easy to adapt to downstream applications, such as the gaming industry, VR, robotics, \etc.

Enhanced by the introduction of the extensive 3D dataset, Objaverse~\cite{deitke2023objaverse, deitke2024objaverse}, numerous 3D reconstruction and generative models emerge.
One notable advancement is LRM~\cite{hong2023lrm}, which pioneers the feed-forward generation of a NeRF model from RGB images.
Subsequent works~\cite{xu2024instantmesh, wei2024mesh, wang2025crm, yang2024hunyuan3d, liu2024meshformer, ge2024prm} extend LRM's NeRF representation to mesh.
However, two key issues persist in these methods.
\textbf{First,} the reliance on RGB images as input leads to significant ambiguity in training.
As shown in Fig.~\ref{fig: ambiguity}(a), the texture often hides the underlying geometry, thereby leading to the conflict of training objectives exhibited in Fig.~\ref{fig: ambiguity}(b).
Furthermore, as demonstrated in Fig.~\ref{fig: ambiguity}(c), RGB images can be rendered from compositions of multiple wrong geometries and textures, driving the network toward an undesirable averaged solution.
\textbf{Second}, most of the existing mesh reconstruction methods employ FlexiCubes~\cite{shen2023flexible} to extract the mesh and utilize differential rasterization for optimization.
However, the Signed Distance Field (SDF) grid defined in FlexiCubes only promises the meaning of positive and negative signs for surface extraction, which makes it difficult to apply 3D supervision.
Moreover, some of its components are redundant for this task, and its regularization losses lead to serious instability in training.

To solve these two challenges, we propose DiMeR, a geometry-texture disentangled feed-forward sparse-view mesh reconstruction model with 3D supervision.
To address the first challenge, training ambiguity, we exploit the inductive bias derived from the \textbf{consistency between normal maps and 3D geometry}.
As shown in Fig.~\ref{fig: ambiguity} (a) and (c), the normal maps consistently align with the surface of the 3D model, offering a more reliable input format for geometry reconstruction.
Building on this inductive bias and the Principle of Occam's Razor~\cite{blumer1987occam}, we disentangle the geometry-texture unified solution space, where multiple solutions correspond to one input, into two individual spaces.
To get the geometry, we \textbf{exclusively} utilize normal maps as the sole input, while utilizing RGB images for obtaining the texture.
Each of them is trained with task-specific supervision: the geometry branch is constrained by normal, depth, and silhouette-mask losses, whereas the texture branch is guided by appearance-based objectives.
Moreover, an accurate geometry should be capable of rendering the light map correctly under arbitrary environmental conditions and various materials.
Therefore, we add the statistical expectation supervision signal by placing the predicted untextured mesh model in multiple environments with randomly assigned materials.
To address the second challenge, limitations remaining in FlexiCubes, we replace the original regularization losses with eikonal loss~\cite{gropp2020implicit} and incorporate 3D supervision via ground-truth mesh.
Furthermore, we simplify the modules with low performance/cost ratios.
Our improved mesh extraction algorithm allows us a higher extraction resolution, compared with other reconstruction models.

Leveraging recent foundation models for normal map prediction~\cite{ye2024stablenormal, he2024lotus}, we can generate highly accurate normal maps from RGB images with minimal error and latency (\textbf{only 200ms}). 
To validate this point and choose the optimal model for DiMeR, we conduct a benchmark evaluation for object-level normal prediction.
To further improve the robustness of DiMeR in practical applications, we introduce noise to the input, normal maps, during training.
\textbf{Equipped with these models, DiMeR also accepts RGB images as raw input, the same as other methods.}

Our DiMeR model is capable of effectively handling various tasks, including sparse-view reconstruction, single-image-to-3D, and text-to-3D.
Extensive experiments demonstrate that DiMeR significantly outperforms previous methods.
Specifically, on the GSO dataset, DiMeR reduces Chamfer Distance by \textbf{22\%}, with an upper bound improvement of \textbf{32\%} when using real normal map inputs.

In general, our contributions can be summarized as follows:
\begin{itemize}
    \item Rethinking the inductive bias for mesh reconstruction, we propose DiMeR, a disentangled framework to train and predict geometry from normal maps and texture from RGB images separately, with decoupled supervision signals.
    \item We enhance the mesh extraction algorithm for this task and introduce the 3D ground truth supervision.
    \item We conduct a benchmark for the foundation models of normal map prediction in object-level tasks.
	\item Numerous experiments demonstrate the superiority and robustness of our DiMeR on reconstruction, single-image-to-3D, and text-to-3D tasks.
\end{itemize}

\section{Related Works}

\subsection{3D Generative Models}

Building upon advancements in 2D diffusion models, DreamFusion~\cite{poole2022dreamfusion} introduced score distillation sampling (SDS) to train 3D representation models like NeRF~\cite{mildenhall2021nerf} and 3DGS~\cite{kerbl20233d} based on text input.
Subsequently, numerous methods have been developed to enhance this approach~\cite{wang2023prolificdreamer, liang2023luciddreamer, chen2023fantasia3d, shi2023mvdream, raj2023dreambooth3d, zhougala3d, chengprogressive3d, bai2023componerf, lin2023magic3d, wang2023score, metzer2023latent, lukoianov2024score, li2025connecting, tang2023dreamgaussian, jiang2024general, yi2023gaussiandreamer}.
However, a significant limitation of these methods is the need to train a separate 3D model for each text input, which can take tens of minutes or even hours per text.
Some approaches attempt to address this by employing SDS to train a feed-forward network~\cite{lorraine2023att3d, jiang2024brightdreamer, li2023instant3d, qian2024atom}, but these are limited to a few specific text subjects, reducing the diversity of the outputs.
Recently, the introduction of large-scale 3D datasets, such as Objaverse~\cite{deitke2023objaverse, deitke2024objaverse}, has enabled models like LRMs~\cite{hong2023lrm, tochilkin2024triposr} to explore feed-forward reconstruction from a single image.
Following this, several methods have been developed to create sparse-view reconstruction models~\cite{tang2025lgm, xu2024grm, zhang2024gs, li2023instant3d2} based on NeRF or 3DGS.
To support real-world applications, leveraging differential marching cube algorithms~\cite{wei2023neumanifold, shen2023flexible}, some methods focus on direct mesh generation~\cite{xu2024instantmesh, wei2024mesh, wang2025crm, liu2024meshformer, ge2024prm}.
Additionally, several 3D diffusion models~\cite{gupta20233dgen, zhang20233dshape2vecset, li2024craftsman, zhang2024clay, ren2024xcube, ren2024scube, zhang2024lagem, xiang2024structured, he2024lucidfusion, szymanowicz2024splatter, lin2025diffsplat} emerge, but they are limited to the generation task and lack strict correspondence with the input image.
Moreover, their inference times range from tens of seconds to several minutes.
Inspired by auto-regressive (AR) models~\cite{tian2024visual, zhou2024transfusion, xie2024show}, some researchers have shifted focus to mesh AR generation~\cite{siddiqui2024meshgpt, chen2024meshanything, chen2024meshanything2, chen2024meshxl, tang2024edgerunner, weng2024pivotmesh, wang2024llama}.
However, these methods typically require the number of mesh faces to be fewer than 6,000, and they exhibit low robustness.
Concurrently, similar to us, Hi3DGen~\cite{ye2025hi3dgen} also found that exclusive utilization of normal maps can enhance the quality of geometry and implemented a diffusion model based on this.

In this paper, we focus on feed-forward sparse-view mesh reconstruction.
Differently, we disentangle the framework into dual branches that predict geometry solely from normal and predict texture from RGB.
To ensure that each branch performs its intended role, we assign branch-specific, unambiguous supervision signals.

\subsection{Multi-view Diffusion Model}

Multi-view diffusion models are designed to generate multi-view images or normal maps from a single image or text prompts, instead of directly producing corresponding 3D models.
This approach is gaining popularity due to the relative simplicity of its task definition, where multi-view images are synthesized first, followed by the use of sparse-view reconstruction models to complete the 3D model generation process.
Zero123~\cite{liu2023zero} introduces explicit control by embedding camera parameters into the conditions of 2D diffusion models.
Following, many methods have achieved significant success to synthesis multi-view images and normal maps~\cite{shi2023zero123++, shi2023mvdream, li2023instant3d2, melas20243d, wang2023imagedream, voleti2024sv3d, wu2024unique3d, li2024era3d, long2024wonder3d, lu2024direct2, lin2025kiss3dgen}.
We employ the image-input 2.5D model, such as zero123++~\cite{shi2023zero123++} and Era3D~\cite{li2024era3d}, to perform the single-image-to-3D task, while we use the text-input 2.5D diffusion model, such as Kiss3DGen, to accomplish the text-to-3D task.
With the continuing progress of such models, DiMeR has the potential to further enhance generation quality.

\subsection{Normal Prediction Foundation Models}

Surface normals precisely describe local surface variation and orientation, making them crucial for 3D reconstruction.
The robustness and accuracy of recent normal prediction foundation models have reached practical levels.
Marigold \cite{ke2024repurposing} first integrates diffusion models into depth and normal estimation, by preserving the prior that the visual generative models learned.
Subsequent works~\cite{fu2024geowizard, bae2024rethinking, ye2024stablenormal, he2024lotus} have further boosted performance while markedly reducing inference latency.
Collectively, these advances provide the possibility for high-quality 3D reconstruction exclusively from predicted normal maps, enabling DiMeR with the RGB image as input.

\begin{figure*}[t!]
    \centering
    \includegraphics[width=1\linewidth]{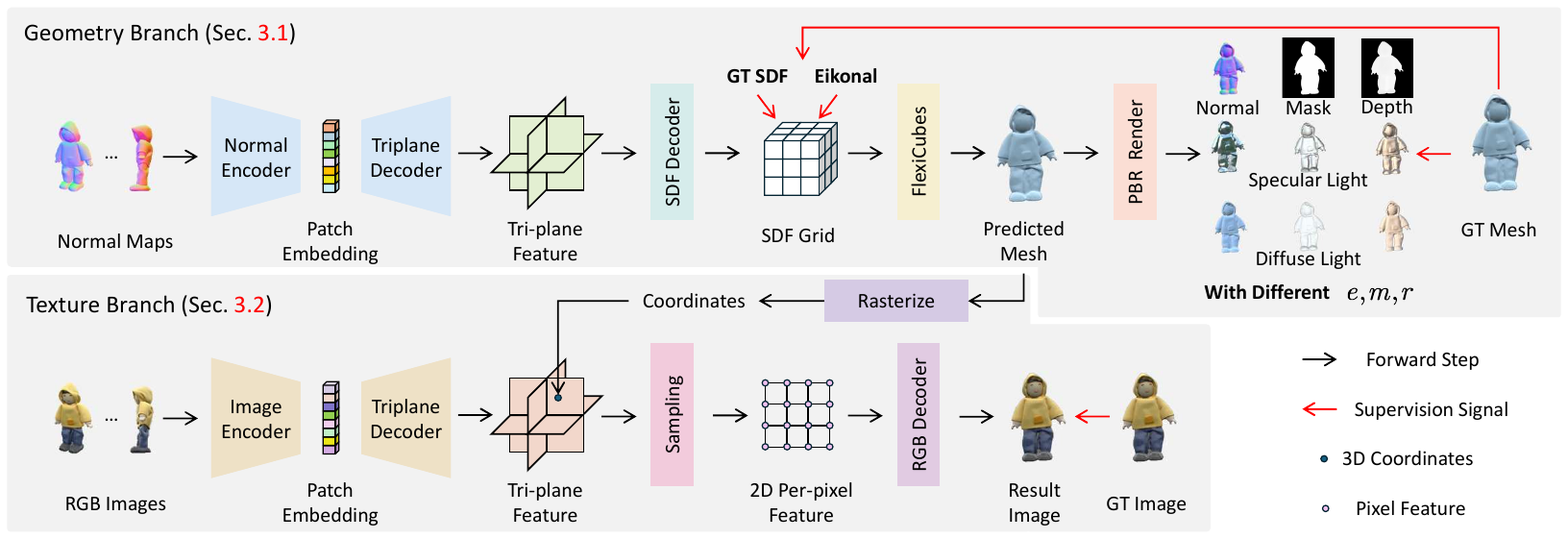}
    \vspace{-20pt}
    \caption{The framework of our DiMeR. The upper part is the geometry branch, and \textbf{exclusively} uses normal maps as input. The lower part is the texture branch.}
    \vspace{-10pt}
    \label{fig: framework}
\end{figure*}

\section{Method}

As shown in Fig.~\ref{fig: framework}, the objective of our DiMeR is to reconstruct the 3D mesh geometry from normal maps and derive texture from RGB images.
We introduce Geometry Branch in Sec.~\ref{sec: geometry}, Texture Branch in Sec.~\ref{sec: texture}, and applications in Sec.~\ref{sec: applacations}.

\subsection{Geometry Branch}
\label{sec: geometry}
As illustrated in Fig.~\ref{fig: ambiguity}, a single RGB image admits many equally plausible solutions in the geometry–texture joint solution space, encouraging the network to learn over-smoothed averages.
Normal maps, in contrast, are uniquely determined by the underlying surface and faithfully encode fine geometric variation.
Guided by Occam’s razor~\cite{blumer1987occam}, we therefore feed only normal maps to the geometry branch, eliminating appearance-induced ambiguities and simplifying the correspondence between input and output.
This design establishes a clearer relationship between the network’s inputs and outputs, ultimately reducing the training complexity.
Supervision is likewise restricted to geometry-specific losses, discarding ambiguous RGB rendering terms.
We further regularize geometry by rendering the untextured mesh with physically based rendering (PBR) under diverse illuminations and materials, matching the resulting lighting maps to statistical expectations.
Finally, we improve the mesh extraction algorithm for greater efficiency and robustness and incorporate direct 3-D supervision.

\noindent \textbf{Network Structure.}
As shown in Fig.~\ref{fig: framework}, the geometry branch of our DiMeR model initiates with normal maps $\mathcal{N} \in \mathbb{R}^{K \times H \times W \times 3}$ of $K$ randomly selected views, alongside their associated camera embeddings $\boldsymbol{\zeta} \in \mathbb{R}^{K \times 16}$.
We opt for a random sampling of input views to improve the model's capability to interpret camera embeddings from arbitrary directions and add slight noise to them, thereby enhancing robustness and reducing dependency on specific input configurations.
Furthermore, this also reduces the requirements for the user input, allowing users to provide inputs from unfixed view directions.
The normal maps $\mathcal{N}$ and their associated camera embeddings $\boldsymbol{\zeta}$ are encoded into patch-wise representations $\mathcal{P}_{g} \in \mathbb{R}^{K \times D \times C}$ using a ViT-based Normal Encoder, where $D$ is the number of patches of each view and $C$ is the dimension of the feature.
Similar to the approach taken by LRM~\cite{wei2024mesh}, we utilize a Triplane Decoder to gather information from the Patch Embedding $\mathcal{P}_{g}$ using several transformer layers~\cite{vaswani2017attention} to synthesize triplane~\cite{chan2022efficient} features $\mathcal{F}_{g} \in \mathbb{R}^{3 \times H^{\prime} \times W^{\prime} \times C_{g}}$.
Subsequently, we extract an SDF grid from the triplane features $\mathcal{F}_{g}$ to apply the differential isosurface construction algorithm, FlexiCubes~\cite{shen2023flexible}, to obtain the vertices and faces for the mesh.
Finally, we can rasterize the mesh to get the normal maps, masks, and depth maps for arbitrary views.
By providing the environment map and assigning different materials (metallic and roughness) to the mesh, we can render the light map (including specular and diffuse) using PBR, for enhancing the supervision from different lighting conditions, which will be introduced in the following part.

\noindent \textbf{Mesh Extraction Algorithm.}
Original FlexiCubes algorithm requires two MLP networks to allow the different weights (each edge and vertex in the grid) and the deformation of the grid.
However, this incurs excessive computational and GPU memory overhead.
Specifically, for a $N^3$ grid, it needs to compute the deformation of $N^3$ vertices and the weight of $12 \times N^3$ edges and $8 \times N^3$ vertices.
Though it is powerful for the tasks of Flexicubes itself, extensive experiments prove that these components contribute disproportionately high computational overhead with minimal performance gains.
As shown in Tab.~\ref{tab: deformation}, we found that removing these networks from the pre-trained model does not adversely affect performance.
Therefore, to enable higher efficient training and higher spatial resolutions, we prune these components to improve computational efficiency and improve the spatial resolution.

\noindent \textbf{Optimization.}
Given the inherent ambiguity introduced by the RGB texture shown in Fig.~\ref{fig: ambiguity}, we exclude RGB loss to enhance training stability.
Consequently, we now exclusively employ geometry-related losses to supervise the geometry branch of our model.

In its original implementation, FlexiCubes incorporates three regularization losses to regularize the SDF grid values generated by the network.
However, extensive experimentation reveals that this approach yielded low stability~\cite{xu2024instantmesh, ge2024prm}.
Furthermore, the design does not produce true SDF representations.
To address these issues, we employ the eikonal loss~\cite{gropp2020implicit} to regularize the whole space as the SDF field, specifically by ensuring the norm of the gradient with respect to the coordinates is normalized to 1.
Nevertheless, computing the derivative for a $N^3$ grid poses significant challenges in terms of computational and GPU memory costs and potential overfitting at specific grid positions.
To mitigate this, we propose randomly sampling positions within the space to compute the eikonal expectation loss, effectively reducing computational demands while maintaining the integrity of the regularization, \ie,
{
	\begin{equation}
		\mathcal{L}_{eik} = \mathbb{E}_{\boldsymbol{x}} (\Vert \nabla_{\boldsymbol{x}} {\text{SDF}(\boldsymbol{x})} \Vert_{2} - 1)^2, \boldsymbol{x} \in \mathbb{R}^{3} \sim Uniform(-1, 1),
        \label{eq: eikonal}
	\end{equation}
}
where we sample $200K$ $\boldsymbol{x}$ in each iteration to calculate the expectation.
Moreover, we use the GT SDF value to supervise the SDF value of grid vertices $\boldsymbol{v} \in \mathbb{R}^{N^3 \times 3}$ in FlexiCubes,
\begin{equation}
	\mathcal{L}_{sdf} = {\Vert \text{SDF}(\boldsymbol{v}) - \text{SDF}_{\text{GT}}(\boldsymbol{v})) \Vert_2}^2.
    \label{eq: sdf}
\end{equation}
To reduce computational overhead, we cache the grid of these SDF values for each object in the training set.

Drawing inspiration from Photometric Stereo~\cite{woodham1980photometric}, we introduce the PBR~\cite{kajiya1986rendering} losses.
The premise is that if the specular and diffuse light maps of a 3D mesh under different environmental lighting conditions and various materials can be accurately rendered in PBR, then the geometry of the predicted mesh model can be deemed correct.
Therefore, we introduce the statistical expectation loss of PBR to supervise the geometry branch,

{
	\setlength{\abovedisplayskip}{1pt}
	\setlength{\belowdisplayskip}{1pt}
	\begin{align}
    \label{eq: spec}
		\mathcal{L}_{spec} = & \ \mathbb{E}_{e, m, r} \left( \text{Spec}(\hat{\mathcal{O}}, e, m, r) - \text{Spec}(\mathcal{O}, e, m, r) \right)^2 \nonumber \\
		& + \text{LPIPS}\left(\text{Spec}(\hat{\mathcal{O}}, e, m, r), \text{Spec}(\mathcal{O}, e, m, r)\right),
	\end{align}
	\begin{align}
    \label{eq: diff}
		\mathcal{L}_{diff} = & \ \mathbb{E}_{e, m, r} \left( \text{Diff}(\hat{\mathcal{O}}, e, m, r) - \text{Diff}(\mathcal{O}, e, m, r) \right)^2 \nonumber \\
		& + \text{LPIPS}\left(\text{Diff}(\hat{\mathcal{O}}, e, m, r), \text{Diff}(\mathcal{O}, e, m, r)\right),
	\end{align}
}
where $\hat{\mathcal{O}}$ is the predicted mesh model, $\mathcal{O}$ is the ground truth mesh model, $e$, $m$, $r$ are the randomly sampled environment, metallic, and roughness, $\text{Spec}(\cdot)$ and $\text{Diff}(\cdot)$ are the rendering functions of specular and diffuse light map, $\text{LPIPS}(\cdot)$ is the perception loss~\cite{zhang2018unreasonable}.
Notably, during the training, we sample different environment, metallic, and roughness to render the light maps for a single object.

We also employ the commonly used normal, depth, and mask losses to supervise the geometry branch.
Specifically,
\begin{equation}
	\mathcal{L}_{nor} = \mathcal{M}_{\text{GT}} \otimes (1 - \hat{\mathcal{N}} \cdot \mathcal{N}_{\text{GT}}),
\end{equation}
\begin{equation}
	\mathcal{L}_{dep} = \mathcal{M}_{\text{GT}} \otimes | \hat{\mathcal{D}} - \mathcal{D}_{\text{GT}} |,
\end{equation}
\begin{equation}
	\mathcal{L}_{mask} = (\hat{\mathcal{M}} - \mathcal{M}_{\text{GT}}) ^ 2,
\end{equation}
where $\otimes$ denotes element-wise production, $\mathcal{M}_{\text{GT}}$ and $\hat{\mathcal{M}}$ are the rendered mask from ground truth mesh model and predicted mesh model, similarly, $\mathcal{N}$ and $\mathcal{D}$ represent normal map and depth map.

In general, the overall loss function is
{
	\begin{equation}
		\mathcal{L}_{g} = \mathcal{L}_{eik} + \mathcal{L}_{sdf} + \mathcal{L}_{spec} + \mathcal{L}_{diff} + \mathcal{L}_{nor} + \mathcal{L}_{dep} + \mathcal{L}_{mask}.
	\end{equation}
}

\subsection{Texture Branch}
\label{sec: texture}

\noindent\textbf{Network Structure.}
As demonstrated in Fig.~\ref{fig: framework}, the texture branch starts from RGB images $\mathcal{I} \in \mathbb{R}^{K \times H \times W \times 3}$ with the camera embeddings $\zeta$.
Similar as the geometry branch, we use a ViT-based Image Encoder to get the Patch Embedding $\mathcal{P}_{c} \in \mathbb{R}^{K \times D \times C}$ and utilize a Triplane Decoder to assemble the information from $\mathcal{P}_{c}$ to get the triplane features $\mathcal{F}_{c} \in \mathbb{R}^{3 \times H^{\prime} \times W^{\prime} \times C_{c}}$ for the texture field representation~\cite{oechsle2019texture}.
Given the predicted shape from the geometry branch, we rasterize the vertex coordinates $\boldsymbol{v}$ into image space,
\begin{equation}
    Coord_{\mathcal{I}} = \text{Rast}(\boldsymbol{v}, \text{Camera}),
\end{equation}
where the pixel value of $Coord_{\mathcal{I}} \in \mathbb{R}^{H \times W \times 3}$ is the global coordinate.
Next, we query the texture feature $\mathcal{F}_{\mathcal{I}} \in \mathbb{R}^{H \times W \times C}$ on triplane $\mathcal{F}_{c}$ for each pixel,
\begin{equation}
    \mathcal{F}_{\mathcal{I}} = \text{Sample}(Coord_{\mathcal{I}}, \mathcal{F}_{c}).
\end{equation}
Finally, we decode the color feature to predict the image
\begin{equation}
    \hat{\mathcal{I}} = \text{RGB\_Decoder}(\mathcal{F}_{\mathcal{I}}).
\end{equation}

\noindent\textbf{Optimization.}
In this branch, we only use RGB loss to supervise the network.
Specifically,
\begin{equation}
    \mathcal{L}_{t} = ( \hat{\mathcal{I}} - \mathcal{I}_{\text{GT}} ) ^ 2 + \text{LPIPS}(\hat{\mathcal{I}}, \mathcal{I}_{\text{GT}}).
\end{equation}

\begin{figure}[t!]
    \centering
    \includegraphics[width=1\linewidth]{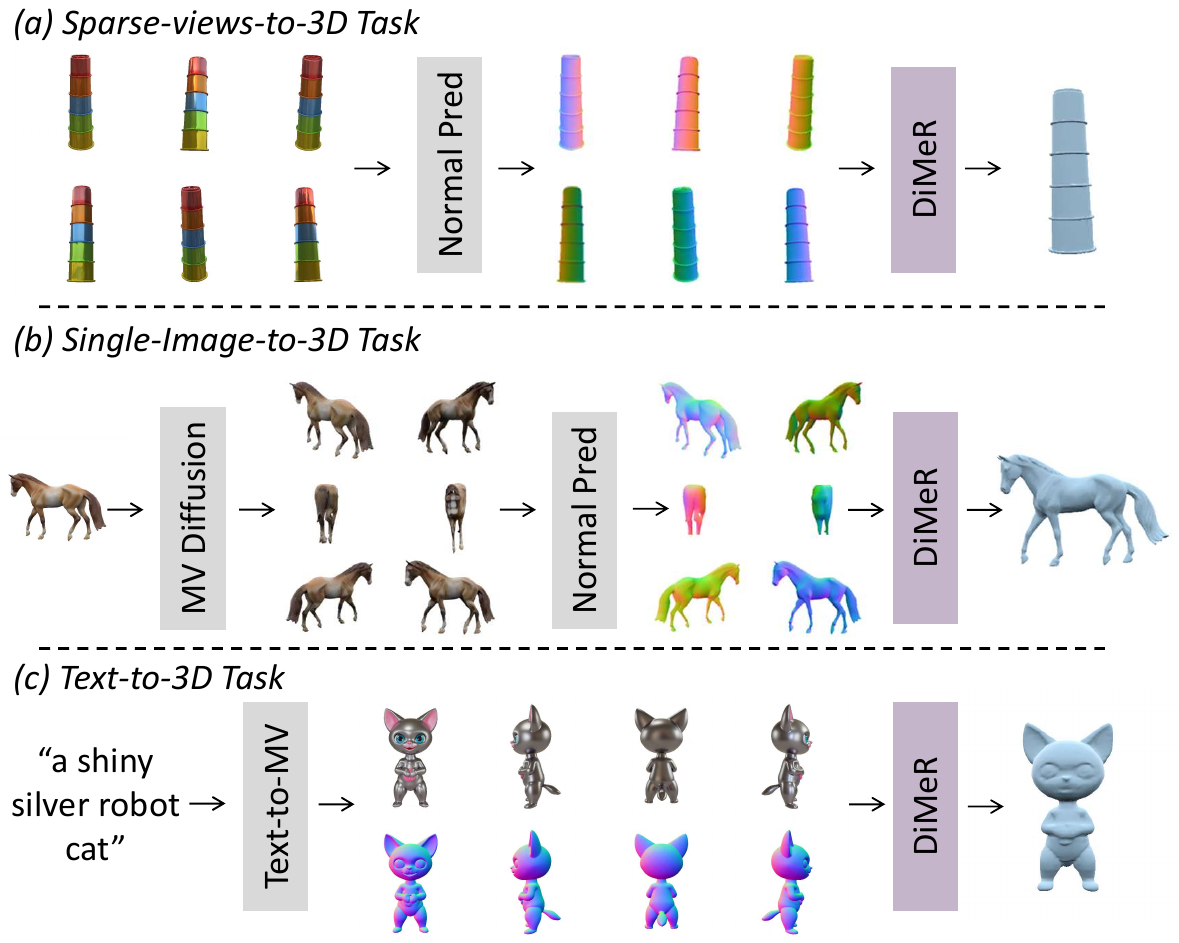}
    \vspace{-20pt}
    \caption{Pipelines for sparse-views, single-image-, and text-to-3D.}
    \vspace{-10pt}
    \label{fig: task_pipeline}
\end{figure}

\subsection{Applications}
\label{sec: applacations}
Besides the sparse-view reconstruction task, our DiMeR is also capable of performing image/text-to-3D tasks.

\noindent \textbf{Single-image-to-3D.}
Given the input image, we first employ Zero-1-2-3++ \cite{shi2023zero123++} or Era3D~\cite{li2024era3d} to generate six images from different viewpoints.
Specifically, the output from zero123++ consists of six views, including the combinations of azimuth and elevation, $(30, 20)$, $(90, \text{-}10)$, $(150, 20)$, $(210, \text{-}10)$, $(270, 20)$, and $(330, \text{-}10)$.
Next, we apply the SoTA normal prediction model Lotus~\cite{he2024lotus} or StableNormal~\cite{ye2024stablenormal} to predict the normal maps for these six views.
Since the predicted normal maps are initially in the local camera coordinate system, we subsequently transform them into the global coordinate system using the transformation matrices corresponding to the six view directions.
Finally, we feed the six transformed normal maps and the RGB images into our DiMeR model to generate the textured mesh.

\noindent \textbf{Text-to-3D.}
This task is approached through two distinct pipelines:
(I) The first pipeline involves using a text-to-image model to generate an RGB image from the input text. Subsequently, we apply the single-image-to-3D pipeline to complete the reconstruction.
(II) With the advancement of diffusion models, Kiss3DGen~\cite{lin2025kiss3dgen} fine-tunes the SoTA text-to-image generative model, FLUX~\cite{blackforest2024flux}, to simultaneously output RGB images along with corresponding normal maps, ensuring high multi-view consistency.
Since our DiMeR supports a dynamic number of input views, we can directly feed the four views from Kiss3DGen into DiMeR for 3D model reconstruction.
The generated high-quality models are presented in Fig.~\ref{fig: teaser} and the supplementary materials.

\begin{table*}[t]
    \centering
    \tabcolsep=0.3cm
    \resizebox{\linewidth}{!}{
    \begin{tabular}{l|ccccc|ccccc}
        \toprule
         Dataset&  \multicolumn{5}{c|}{GSO}&  \multicolumn{5}{c}{OmniObject3D}\\ \midrule
           Metric&  CD ($\downarrow$)&  F1 ($\uparrow$)&  PSNR ($\uparrow$)& SSIM ($\uparrow$) &  LPIPS ($\downarrow$)&  CD ($\downarrow$)&  F1 ($\uparrow$)&  PSNR ($\uparrow$) & SSIM ($\uparrow$) & LPIPS ($\downarrow$)\\ \midrule
           InstantMesh&  0.045&  0.964& 18.51& 0.846&  0.150& 0.039 & 0.983 & 18.44& 0.842& 0.153\\
           PRM&  0.041&  0.977& 21.68& 0.869& 0.126& \third 0.034& \third 0.991&  21.65& 0.865&0.135\\ 
          DiMeR~(\textcolor{blue}{GT})& \first 0.028& \first 0.992& \first 23.40& \first 0.883& \first 0.095& \first 0.024 & \first 0.996 & \first 23.04& \first 0.871&\first 0.112\\
         \shaddow $\Delta$&  \shaddow 31.7\% $\downarrow$& \shaddow +0.015 & \shaddow +1.72 & \shaddow +0.014&\shaddow 24.6\% $\downarrow$&\shaddow 29.4\% $\downarrow$ &\shaddow +0.005 &\shaddow +1.39&\shaddow +0.006&\shaddow 17.0\% $\downarrow$\\ \midrule
           DiMeR~(Lotus)&\third 0.033 &\second0.988&\third  22.57&\third   0.874&\second  0.103&\third  0.034 &  0.989&\third  21.88&\second  0.866&\third  0.126\\
           DiMeR~(SN)&\second 0.032 &\second 0.988 &\second 22.89&\second 0.875&\second 0.103&\second 0.030 &\second 0.993 &\second 22.15&\third 0.865&\second 0.121\\

           \bottomrule
    \end{tabular}}
    \caption{Quantitative results for reconstruction task. CD means Chamfer Distance. DiMeR (Lotus) and DiMeR (SN) are the reconstruction results from the normal map predicted by Lotus~\cite{he2024lotus} and  StableNormal~\cite{ye2024stablenormal} separately. DiMeR (GT) is from the ground truth normal. \colorbox[HTML]{F4B5B4}{value} means first-best, \colorbox[HTML]{F9DAB7}{value} means second-best, \colorbox[HTML]{FFFFBB}{value} means third-best.}
    \vspace{-20pt}
    \label{tab: reconstruction}
\end{table*}

\section{Experiment}

\subsection{Implementation Details}
\label{sec: implementation}
\noindent \textbf{Datasets.}
We train DiMeR with the filtered Objaverse \cite{deitke2023objaverse} according to the mesh quality, in a total of $98,526$ objects.
For test datasets, we choose the widely used GSO~\cite{downs2022google} and OmniObject3D~\cite{wu2023omniobject3d}.
We use all $1,029$ objects in GSO and randomly select 5 objects for each class in OmniObject3D.

\noindent \textbf{Evaluation Protocol.}
For 3D metrics, we sample $32,000$ points on the surface to compute commonly used Chamfer Distance (CD) and F1-Score@0.1 to evaluate the quality of geometry.
For 2D metrics, we compute the PSNR, SSIM, and LPIPS to evaluate the rendering quality over $8$ rendered views.
We rescale and align the generated meshes and ground truth meshes for fair comparison.

\noindent \textbf{Training.}
We set the total batch size to $64$, with learning rate of $4 \times 10^{-6}$ for geometry branch and $4 \times 10^{-5}$ for texture branch.
The resolution of the triplane is $3 \times 64 \times 64$, and the SDF grid is $192^3$, which is higher than baselines benefiting from our enhancement of mesh extraction methods.
The resolutions of input and supervision are $512 \times 512$.
For PBR statistical expectation loss, we place the predicted meshes in $10$ different lighting environments and apply $10$ different materials, rendering from different $10$ views during training.
We train the geometry branch for two days and the texture branch for one day on $8$ H100 GPUs.

\subsection{Quantitative Comparison}
\label{sec: quantitative}

\begin{table}[t]
    \centering
    \tabcolsep=0.4cm
    \resizebox{\linewidth}{!}{
    \begin{tabular}{l|cc|cc}
        \toprule
         Dataset&  \multicolumn{2}{c|}{GSO}&  \multicolumn{2}{c}{OmniObject3D}\\ \midrule
           Metric&  CD ($\downarrow$)&  F1 ($\uparrow$)&  CD ($\downarrow$)&  F1 ($\uparrow$)\\ \midrule
           CRM&  0.144&  0.781&  0.114& 0.854\\
           InstantMesh& \third 0.066& \third 0.950& \third 0.074& \third 0.937\\
           PRM& \second 0.059& \second 0.961&  \second 0.064&   \second 0.957\\
           Trellis& 0.119 & 0.859 &  0.090&   0.902\\
           DiMeR& \first 0.052& \first 0.981&  \first 0.060&   \first 0.964\\
           \bottomrule
    \end{tabular}}
    \caption{Single-image-to-3D task. All the methods use the same single image input. Our DiMeR is equipped with Stable-Zero123++~\cite{shi2023zero123++} and StableNormal~\cite{ye2024stablenormal}.}
    \vspace{-20pt}
    \label{tab: single-image}
\end{table}

\noindent \textbf{Reconstruction Task.}
As shown in Tab.~\ref{tab: reconstruction}, we compare our DiMeR on the sparse-view reconstruction task using the same 6 randomly sampled input views.
Since some sparse-view reconstruction methods, like CRM~\cite{wang2025crm}, are limited to only support specific views (six orthogonal views), we compare them in single-image-to-3D tasks.
Additionally, because MeshFormer~\cite{liu2024meshformer} is not open-source work, we are unable to perform an accurate quantitative comparison.
Therefore, we only provide qualitative visual comparisons.
For the comparison, we select state-of-the-art (SoTA) methods that are accessible, including InstantMesh~\cite{xu2024instantmesh} and PRM~\cite{ge2024prm}.
Experiments show that our method can surpass the SoTA methods by a large margin, whatever using GT (31.7\% gain) or predicted normal maps (22.0\% gain) from StableNormal-Turbo~\cite{ye2024stablenormal}.
\textbf{Notably, when equipped with normal map prediction models, the input to DiMeR remains the same as the baselines, relying solely on RGB images}.
Furthermore, following the improvement of normal prediction models,  there is still room for continued improvement in the performance of DiMeR.

\noindent \textbf{Sinle-Image-to-3D Task.}
As demonstrated in Tab.~\ref{tab: single-image}, we compare our DiMeR with CRM~\cite{wang2025crm}, InstantMesh~\cite{xu2024instantmesh}, PRM~\cite{ge2024prm}, and Trellis~\cite{xiang2024structured} using same single image input.
Our pipeline for this task is shown in Fig.~\ref{fig: task_pipeline}(b), where we use Lotus~\cite{he2024lotus} to predict normal maps from the output of zero123++~\cite{shi2023zero123++}.
Since the single-image-to-3D problem is inherently ill-posed, the unseen portions of the data cannot be accurately inferred from a single image alone. 
Consequently, we select 500 relatively clear data points for meaningful and valuable evaluation.
Notably, while Trellis produces high-quality mesh generation, issues with consistency in the input image persist. This inconsistency can be attributed to Trellis' generative nature rather than being a deterministic model.
This point is also highlighted in the qualitative comparisons in Fig.~\ref{fig: image-to-3D}.
In contrast, the reconstruction models, such as our DiMeR, PRM, and InstantMesh, have the advantages for the accurate alignment with input image based on the prediction of zero123++.

\subsection{Qualitative Comparison}
\label{sec: qualitative}

\begin{figure*}[t!]
    \centering
    \includegraphics[width=1\linewidth]{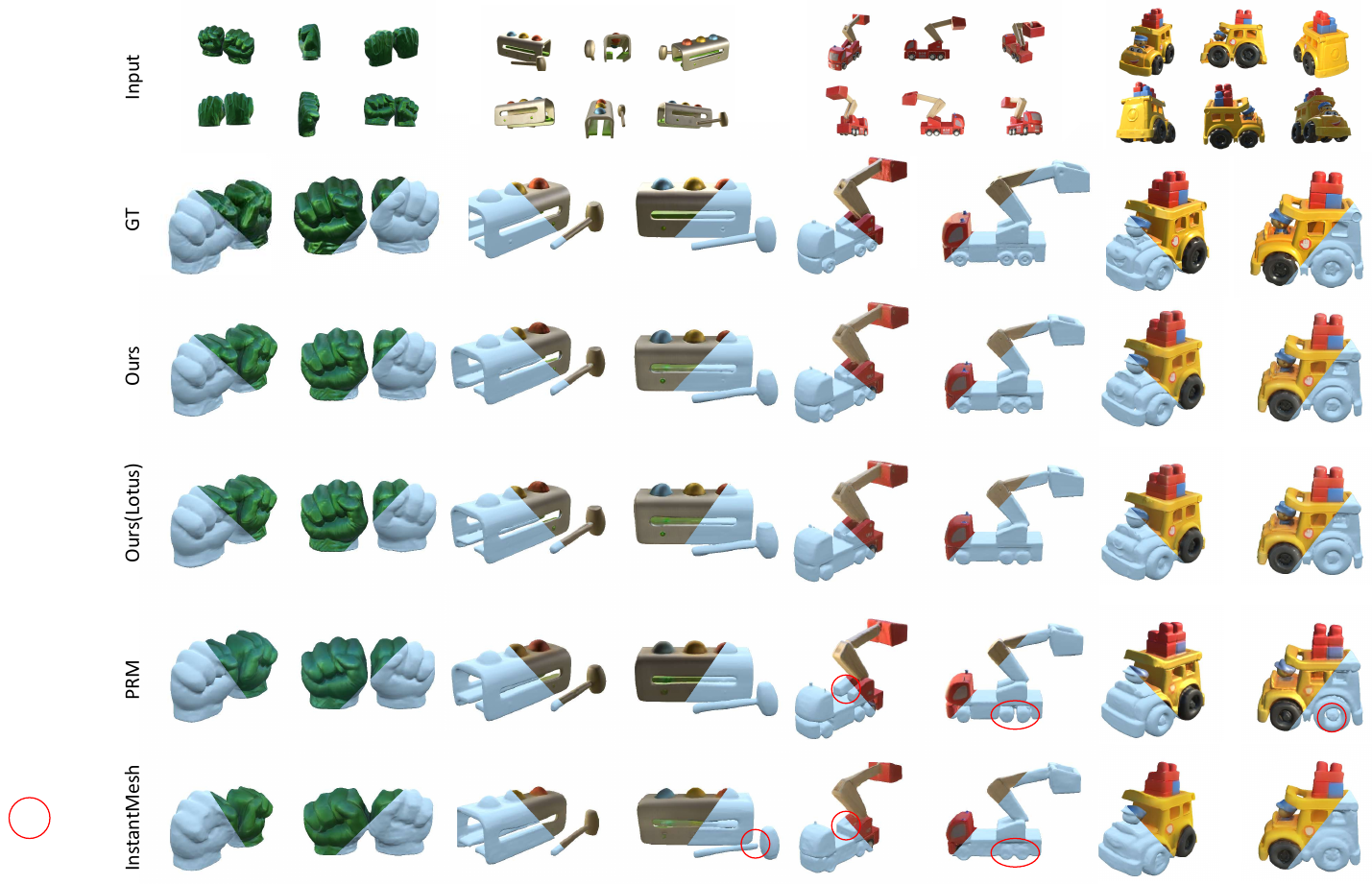}
    \vspace{-20pt}
    \caption{The qualitative comparison for sparse view reconstruction.}
    \vspace{10pt}
    \label{fig: reconstruction}
\end{figure*}

\noindent \textbf{Reconstruction Task.}
As demonstrated in Fig.~\ref{fig: reconstruction}, we present a visual qualitative comparison of various methods. A comparison between the rows labeled "Ours" and "Ours (Lotus)" shows similar performance, highlighting that normal prediction models effectively support DiMeR. This suggests that DiMeR, when combined with such models, is capable of surpassing previous methods in realistic applications.
Furthermore, DiMeR outperforms previous mesh reconstruction models, such as InstantMesh and PRM, in terms of reconstructing finer details.

\begin{figure*}[t!]
    \centering
    \includegraphics[width=1\linewidth]{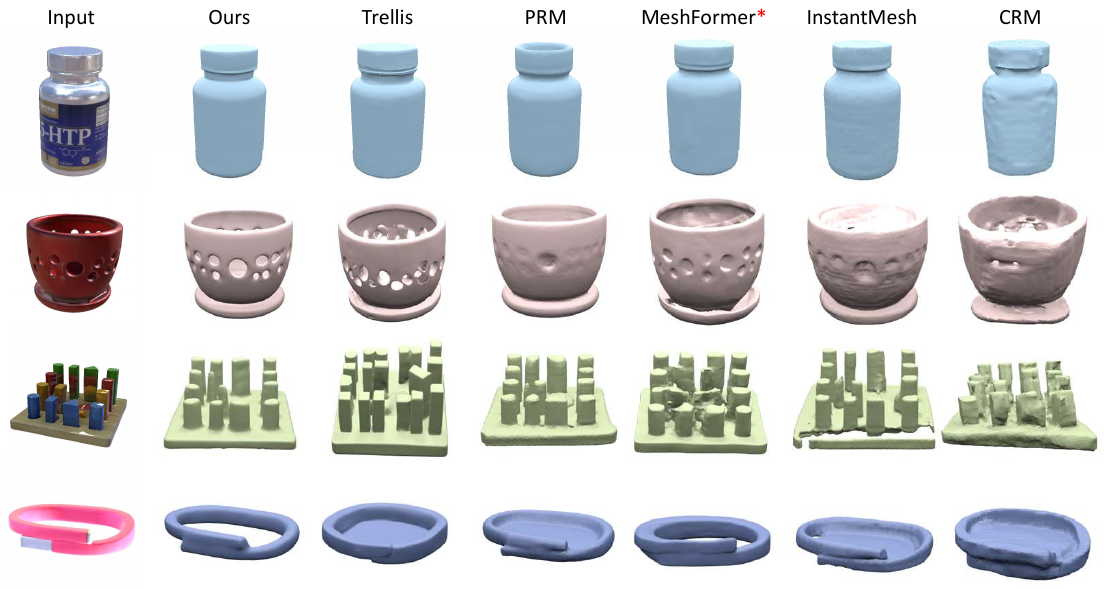}
    \vspace{-20pt}
    \caption{The qualitative comparison for single-image-to-3D. Please note that the results of MeshFormer are obtained from their project page and do not use the same input as other methods.}
    \vspace{20pt}
    \label{fig: image-to-3D}
\end{figure*}

\noindent \textbf{Single-image-to-3D.}
As shown in Fig.~\ref{fig: image-to-3D}, we compare our method with SoTA methods including Trellis~\cite{xiang2024structured}, PRM~\cite{ge2024prm}, MeshFormer~\cite{liu2024meshformer}, InstantMesh~\cite{xu2024instantmesh} and CRM~\cite{wang2025crm}.
Notably, since the 3D results for MeshFormer are only available from their project page and the corresponding input images are not provided, we are unable to conduct a direct comparison using the same input. 
In contrast, the other methods use the same input images for comparison.
Due to the inherent characteristics of the generative diffusion model, Trellis often generates 3D mesh models that exhibit inconsistencies with the input images, although it maintains high quality.
Specifically, the cup's holes in the second column and the number of pillars in the third column are mismatched.
Moreover, the other methods encounter difficulties in generating holes and rings accurately.
In summary, our DiMeR achieves the best consistency and quality.

\begin{table*}[t]
    \centering
    \tabcolsep=0.25cm
    \resizebox{\linewidth}{!}{
    \begin{tabular}{l|ccccc|ccccc|c}
        \toprule
         Dataset&  \multicolumn{5}{c|}{GSO}&  \multicolumn{5}{c|}{OmniObject3D} & - \\ \midrule
           Metric&  mean ($\downarrow$)& median ($\downarrow$)&  $11.25^{\circ}$ ($\uparrow$) &  $22.5^{\circ}$ ($\uparrow$)&  $30^{\circ}$ ($\uparrow$)& mean ($\downarrow$)& median ($\downarrow$)&  $11.25^{\circ}$ ($\uparrow$) &  $22.5^{\circ}$ ($\uparrow$)&  $30^{\circ}$ ($\uparrow$)& Latency ($\downarrow$)\\ \midrule
           GeoWizard&  17.673&  14.307&  \first 48.309& 74.908&83.097&23.129&20.156&28.272&61.215&74.122&2102 ms\\
 Marigold& 17.400& \third  14.305& \second 47.303& \third 76.058&84.197&22.934&20.243&28.867&61.201&73.824&260 ms\\
           DSINE& 17.953& 14.857& 45.543& 72.951&82.535&23.010&20.116&\third 29.182&\third 62.140&75.082&\first 59 ms\\
           Lotus-G& \third 
 17.151& \second  13.920&  45.343&   \second 76.831&\third 85.277&\third 21.523&\second 19.048&\second 30.836&\second 64.828&\third 77.359&\second 130 ms\\
           SN V1.8.1& \second 16.818& 14.743&  39.860&   74.524&\first 86.424&\second 21.205&\third 19.468&25.677&61.917&\second 77.916&\third 236 ms\\
           Lotus-D& \first 16.606& \first 13.377&  \third 47.218&   \first 78.076&\second 86.166&\first 21.065&\first 18.622&\first 32.118&\first 66.216&\first 77.968&\second 130 ms\\
           \bottomrule
    \end{tabular}}
    \caption{Benchmark for normal map prediction of foundation models on object scenario. The latency is evaluated on a single A800 GPU.}
    \vspace{-15pt}
    \label{tab: normal benchmark}
\end{table*}

\subsection{Benchmark for Normal Prediction Foundation Models}
\label{sec: normal benchmark}

\begin{figure*}[t!]
    \centering
    \includegraphics[width=1\linewidth]{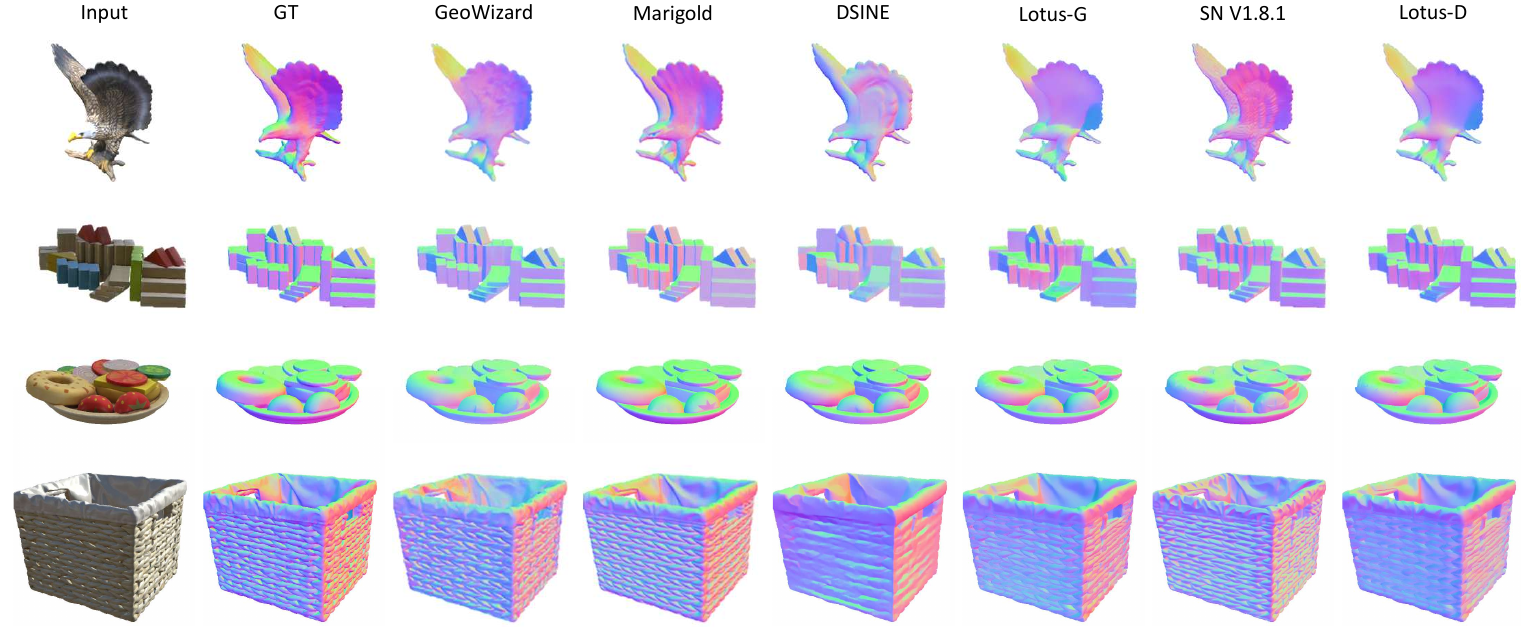}
    \vspace{-20pt}
    \caption{The qualitative comparison for normal prediction foundation models.}
    \label{fig: benchmark}
\end{figure*}

To determine whether recent normal-prediction foundation models meet the quality requirements of our pipeline, we evaluate Lotus~\cite{he2024lotus}, StableNormal~\cite{ye2024stablenormal}, DSINE~\cite{bae2024rethinking}, Marigold~\cite{ke2024repurposing}, and GeoWizard~\cite{fu2024geowizard} on the GSO~\cite{downs2022google} and OmniObject3D~\cite{wu2023omniobject3d} benchmarks.
For each object, six randomly sampled views are rendered, producing paired RGB images, masks, and ground-truth normal maps.
Using the RGB inputs, we measure mean and median angular error, the proportion of pixels with error below $11.25^{\circ}$, $22.5^{\circ}$, and $30^{\circ}$, and inference time.
When computing these metrics, we only use the foreground pixels.
As summarised in Tab.~\ref{tab: normal benchmark}, StableNormal and Lotus offer the best balance of accuracy and speed, adding only negligible latency.
Correspondingly, as demonstrated in Tab.~\ref{tab: reconstruction}, even with errors, \textbf{our DiMeR still outperforms previous methods by a large margin, accepting the same RGB input}.
Among the reported metrics, the mean error and the fraction of pixels with error below the threshold $30^{\circ}$ are most indicative of prediction stability.
Large errors markedly impact reconstruction quality.
Ongoing advances in normal-prediction models are therefore expected to further improve DiMeR’s performance.
We also provide the qualitative comparison in Fig.~\ref{fig: benchmark}.

\subsection{Ablation Studies}
\label{sec: ablation}

In this section, we validate our key designs.
All experiments are conducted based on the GSO dataset.

\begin{table}[t]
\centering
\begin{minipage}[t]{0.55\linewidth}
\centering
\resizebox{\linewidth}{!}{
\begin{tabular}{l|ccc}
\toprule
Input      & RGB & RGB+Normal & Normal \\ \midrule
CD        &    0.041&  0.041&  0.028\\
F1 &    0.971&  0.981&  0.992\\ 
\bottomrule
\end{tabular}}
\caption{The ablation studies of different input formats.}
\label{tab: input}
\end{minipage} \hfill
\begin{minipage}[t]{0.42\linewidth}
\centering

\resizebox{\linewidth}{!}{
\begin{tabular}{l|ccc}
\toprule
Method      & FlexiCubes & Ours \\ \midrule
CD        &  0.037  &  0.028 \\
F1 &  0.975  &  0.992 \\ 
\bottomrule
\end{tabular}}
\caption{The ablation studies of regularization losses.}
\label{tab: regularization}
\end{minipage}
\end{table}

\begin{table}[t]
\vspace{-20pt}
\centering
\begin{minipage}[t]{0.31\linewidth}
\centering
\resizebox{\linewidth}{!}{
\begin{tabular}{l|cc}
\toprule
    & w/o & w/  \\ \midrule
CD &  0.039  & 0.028 \\
F1 &  0.973  & 0.992 \\ 
\bottomrule
\end{tabular}}
\caption{The ablation studies of PBR expectation losses.}
\label{tab: pbr losses}
\end{minipage} \hfill
\begin{minipage}[t]{0.65\linewidth}
\centering
\resizebox{\linewidth}{!}{
\begin{tabular}{l|cc|cc}
\toprule
Method      & CD & F1 & GPU Mem & Infer \\ \midrule
w/ & 0.045   &  0.964  & 73GB & 0.5s \\
w/o & 0.045   & 0.963 & 48GB & 0.2s \\ 
\bottomrule
\end{tabular}}
\caption{The ablation studies of the effectiveness of Deformation and Weight MLP. GPU Mem is training occupancy.}
\label{tab: deformation}
\end{minipage}
\vspace{-25pt}
\end{table}

\noindent \textbf{Input Disentanglement.}
As shown in Tab.~\ref{tab: input}, we compare the performance of DiMeR using different input formats. The comparison between the first two columns, ``RGB'' and ``RGB+Normal'', illustrates that incorporating geometry information results in a slight improvement in effectiveness. Furthermore, the third column, labeled ``Normal'', demonstrates a significant performance gain over ``RGB+Normal''. This improvement underscores the strong inductive bias between normal maps and 3D geometry. Additionally, since the input is encoded as patch embeddings, using mixed input produces more patches, resulting in increased GPU memory usage and computational overhead. In contrast, the exclusive use of normal maps maintains the same resource consumption as RGB inputs.

\noindent \textbf{Regularization Loss.}
As demonstrated in Tab.~\ref{tab: regularization}, we show that Eq.~\ref{eq: eikonal} and Eq.~\ref{eq: sdf} can replace the original loss functions used in FlexiCubes, performing improved performance. With the regularization loss employed in FlexiCubes, the training process becomes unstable and struggles to proceed beyond 10,000 iterations, resulting in unsatisfactory network convergence. By introducing the eikonal loss and incorporating 3D ground truth, we stabilize the training process, achieving significantly better performance.

\noindent \textbf{PBR Loss.}
As shown in Tab.~\ref{tab: pbr losses}, we validate the effectiveness of the PBR expectation loss (Eq.~\ref{eq: spec} and Eq.~\ref{eq: diff}). If the lighting map can be accurately computed under varying environmental lighting conditions and across different materials, we can conclude that the predicted mesh aligns well with the ground truth mesh. To achieve this, we assign different materials to the single predicted mesh and place it in various environments. The introduction of PBR losses leads to significant improvements.

\noindent \textbf{Deformation and Weight MLP in FlexiCubes.}
As shown in Tab.~\ref{tab: deformation}, we demonstrate that the improvements gained from the deformation and weight MLP are not worthy enough compared with their computational cost. The experiments are conducted using the official pretrained weights of InstantMesh, and similar experiments based on PRM are provided in the supplementary material. Upon removing the deformation network and weight network from FlexiCubes, we observe minimal impact on inference performance, almost no decrease. However, these two networks significantly increase computational workload (about $ 2.5 \times $ computation overhead) and GPU memory consumption (about $ 1.5 \times $ GPU memory occupancy in training). Consequently, we opt to exclude them from DiMeR in order to improve the spatial resolution.

\section{Conclusion}
In this paper, we propose DiMeR, a disentangled dual-stream framework with 3D supervision for feed-forward sparse-view mesh reconstruction.
By driving the geometry branch exclusively with normal maps and leaving RGB information to a separate texture branch, DiMeR clearly separates conflicting objectives and grounds training on unambiguous supervision signals.
To enhance the training effectiveness and spatial resolution, DiMeR improves the mesh extraction algorithm by redesigning the regularization losses, introducing 3D ground-truth supervision, and removing redundant modules.
Extensive experiments confirm that DiMeR surpasses state-of-the-art baselines across multiple tasks, such as sparse-view-to-3D, image-to-3D, and text-to-3D, highlighting both its effectiveness and robustness. As normal-prediction models continue to improve, DiMeR’s performance is likely to advance further.


\clearpage

\begin{figure*}[t!]
    \centering
    \includegraphics[width=1\linewidth]{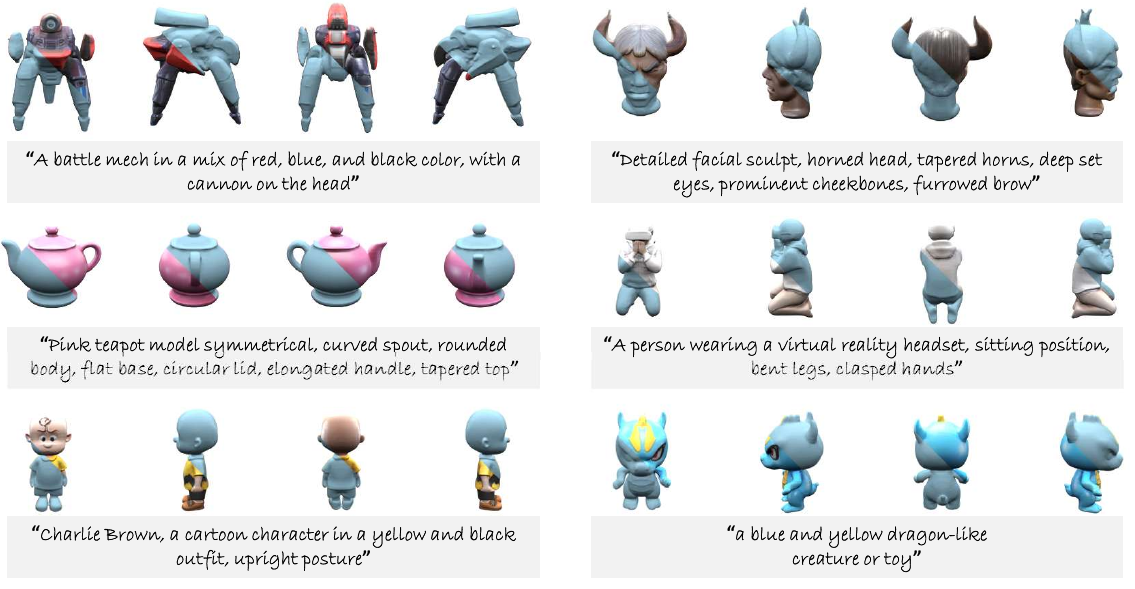}
    \vspace{-20pt}
    \caption{The generation results for text-to-3D.}
    \label{fig: text-to-3D}
\end{figure*}

\clearpage

\bibliographystyle{ACM-Reference-Format}
\bibliography{sample-bibliography}

\clearpage


\end{document}